\documentclass[conference]{IEEEtran}
\IEEEoverridecommandlockouts
\usepackage{cite}
\usepackage{amsmath,amssymb,amsfonts}
\usepackage{algorithmic}
\usepackage{graphicx}
\usepackage{textcomp}
\usepackage{xcolor}
\usepackage{comment}
\usepackage{array}
\usepackage{tabu}
\usepackage{booktabs}
\usepackage{multirow}
\usepackage{lipsum}
\usepackage[bookmarks=false]{hyperref}
\usepackage{amssymb}
\usepackage{pifont}
\usepackage{multicol}
\usepackage{pbox}

\newcommand{\cmark}{\ding{51}}
\newcommand{\xmark}{\ding{55}}

\newcommand{\mytabheader}[1]{\textbf{\textit{#1}}}

\def\BibTeX{{\rm B\kern-.05em{\sc i\kern-.025em b}\kern-.08em
    T\kern-.1667em\lower.7ex\hbox{E}\kern-.125emX}}
\begin{document}

% (?P<bef>[^\s])\s\s(?P<aft>[^\s])

\title{Spotting insects from satellites: modeling the presence of \textit{Culicoides imicola} through Deep CNNs}

\author{

\IEEEauthorblockN{
Stefano Vincenzi\IEEEauthorrefmark{1}, \ Angelo Porrello\IEEEauthorrefmark{1}, \ Pietro Buzzega\IEEEauthorrefmark{1},
Annamaria Conte\IEEEauthorrefmark{2},\ Carla Ippoliti\IEEEauthorrefmark{2},\ Luca Candeloro\IEEEauthorrefmark{2},\\ Alessio Di Lorenzo\IEEEauthorrefmark{2}, Andrea Capobianco Dondona\IEEEauthorrefmark{2}, \ Simone Calderara\IEEEauthorrefmark{1}}
\\

\IEEEauthorblockA{ 
\IEEEauthorrefmark{1}AImageLab, University of Modena and Reggio Emilia, Modena, Italy\\
\IEEEauthorrefmark{2}Istituto Zooprofilattico Sperimentale dell'Abruzzo e del Molise `G.Caporale', Teramo, Italy}
%\IEEEauthorrefmark{3}Farm4Trade Srl, Chieti, Italy \\
}

\maketitle

\begin{abstract}
Nowadays, Vector-Borne Diseases (VBDs) raise a severe threat for public health, accounting for a considerable amount of human illnesses. Recently, several surveillance plans have been put in place for limiting the spread of such diseases, typically involving on-field measurements. Such a systematic and effective plan still misses, due to the high costs and efforts required for implementing it. Ideally, any attempt in this field should consider the triangle vectors-host-pathogen, which is strictly linked to the environmental and climatic conditions. In this paper, we exploit satellite imagery from Sentinel-2 mission, as we believe they encode the environmental factors responsible for the vector's spread. Our analysis - conducted in a data-driver fashion - couples spectral images with ground-truth information on the abundance of \textit{Culicoides imicola}. In this respect, we frame our task as a binary classification problem, underpinning Convolutional Neural Networks (CNNs) as being able to learn useful representation from multi-band images. Additionally, we provide a multi-instance variant, aimed at extracting temporal patterns from a short sequence of spectral images. Experiments show promising results, providing the foundations for novel supportive tools, which could depict where surveillance and prevention measures could be prioritized.
\end{abstract}


\section{Introduction}

Recent incursions into Europe of some vector-borne diseases (Chikungunya and Malaria), the persistence and spread of others (West Nile Disease, Bluetongue and Lyme disease), as well as the risk of introduction of new exotic diseases (Rift Valley Fever, Crimean Congo Hemorrhagic Fever, African Horse Sickness) prompt an integrated surveillance action. On the one hand, such a solution may identify both the presence and the abundance of the vectors (e.g., ticks, culicoides, mosquitoes); on the other hand, it may lead to a deeper knowledge of the territory and the potential habitats of different arthropod species. Over the years, public institutions and organizations implemented various surveillance plans for monitoring the spread of vectors in Europe~\cite{semenza2017vector}. However, a systematic and intensive plan would require extremely high resources and funds. In this respect, the identification of specific monitoring areas could bring to a cost reduction and fine-grained entomological surveillance strategies.

The survival and spread of vectors and pathogens in a given area rely on several factors, such as climate, vegetation, landscape composition, and hosts' availability. In the recent past, various studies relate the presence and the abundance of vectors with different climatic and environmental factors. As an example, in~\cite{conte2007influence}, the authors defined a series of liable variables, such as temperature, aridity index, elevation, and vegetation index. The analysis of these factors capture the distribution of different species of \textit{Culicoides} and the environment favorable to them. Furthermore, the authors of~\cite{mcintyre2017systematic} pointed out the climate change as responsible for the spread of certain pathogens, which in turn are involved in the spread of the above-mentioned diseases.

To sum up, the environmental and climatic conditions represent a crucial aspect for modeling the distribution and abundance of vectors. Nowadays, dozens of satellites provide images at very high resolutions and with unprecedented rhythms. In this work, we explore the analysis of these images for predicting the presence of the \textit{Culicoides imicola}, the latter being among the species responsible for the Bluetongue disease. In more details, we conduct all the analyses on the satellite imagery coming from the Sentinel 2 mission, handled by the European Space Agency (ESA)~\cite{drusch2012sentinel,berger2012esa} within the Copernicus program. Copernicus builds on a global network of thousands of sensors for land, ocean and atmospheric monitoring of Earth, resulting in an impressive number of daily observations. The Sentinel-2A and Sentinel-2B satellites are equipped with a multi-spectral device (MSI) capable of acquiring 13 spectral bands, ranging from visible (RGB) to short-wave infrared (SWIR), including near-infrared (NIR).

These multi-band images could provide an enormous amount of information about the landscape composition, fragmentation and configuration, which relate to the presence and abundance of vectors. Aiming to capture these kind of relationships among data, we gathered a new dataset, which pairs spectral bands coming from the Sentinel mission with ground-truth measures of presence/absence for the \textit{Culicoides imicola} in a certain area.

However, analyzing these data with shallow machine learning models would require intense efforts during the features extraction stage, which could be enhanced using deep learning techniques. These techniques have already been used in various researches related to the remote sensing field, such as soil and crop classification. In our work, we frame the problem of estimating the presence of \textit{C. imicola} as a binary classification task, and take advantage of Deep Convolutional Neural Networks (DCNNs)~\cite{sutskever2012imagenet}, which provide meaningful and hierarchical representations from multi-band images. To the best of our knowledge, their application in the field of vector-borne diseases (VBDs) is still very limited.

We evaluate two kinds of approaches: on the one side, we adopt a single multi-band image for predicting the given binary target; on the other side, we underpin a sliding window approach over time, where the network takes as input a sequence of multi-band images. Those are acquired during the month preceding the field catch of the vector: this way, we aim to assess whether a temporal series of inputs could be more informative. 
In the above-mentioned approaches, we rely on ResNet~\cite{he2016deep} as a backbone network for the feature extraction stage, coupled with an aggregation module for merging representations within the input window. For assessing the merits of different inputs, we test our solution with: i) RGB channels ii) the NDVI and NDWI indices; iii) all the 13 raw bands captured by the Sentinel satellites. Briefly, experiments show two insightful remarks: firstly, the spectral bands encode important features for the task at hand; secondly, the multi-instance approach outperforms the single-instance one.

The manuscript is organized as follows: related works are discussed in Section \ref{sec:related}. We describe the datasets we used in Section \ref{sec:dataset}. In Section \ref{sec:models}, we introduce the proposed methods. We report the experiments and ablation studies we conducted in Section \ref{sec:experiments}. We finally draw several conclusions in Section \ref{sec:conclusion}. 

\section{Related Works}
\label{sec:related}

\subsection{Vector-borne diseases and machine learning approaches}

Recently, several studies exploit machine learning algorithms for analyzing the spread of VBDs in a given area. In this respect, the authors of~\cite{ippoliti2019defining} took advantage of eco-regions, namely areas' within which there are associations of interacting biotic and abiotic features'~\cite{bailey2004identifying}. The underlying idea relies on the observation that areas with similar conditions are potentially prone to similar disease risk, even though they are distant from each other. Basically, they proposed a map of eco-climatic Italian regions, the latter identified through a data-driven approach. Formally, their clustering method relies on seven environmental and climatic variables, relevant for the spread of various VBDs. Afterward, they related the resulting clustering map with ground-truth information about the Bluetongue vectors and West Nile Disease (WND) outbreaks in Italy. They showed that the habitat of \textit{C. imicola} was represented by two out of the twenty-two eco-regions, thus indicating high-risk areas and therefore where surveillance measures could be prioritized. 

Differently,~\cite{ducheyne2013abundance} predicted both the presence and abundance of five different vectors (\textit{C. imicola}, \textit{C. newsteadii}, \textit{Pulicaris complex}, \textit{C. punctatus} and \textit{C. obsoletus}) across the Spanish territory. The authors pointed out the abundance as an important factor, essential for mathematical models for estimating both the potential of the transmitted disease and its future behaviour. Previous approaches considered the abundance linearly related to the probability of presence. Instead, they compute the former by combining the predicted probability with other features, such as temperature or precipitations. To this aim, they used various dataset (e.g. GTOPO30, CORINE etc.) providing meteorological and environmental features (e.g. land cover, vegetation, ground water etc.), mainly derived from satellite imagery. Similarly to us, they gather ground-truth information from traps located near the farms.

In technical details, they fed these data to an ensemble model, exploiting a classification and regression forest for estimating the presence and the abundance respectively. Results show particular factors being fundamental for the spread of specific vectors. In particular, focusing on the \textit{C. imicola}, the authors found summer rainfall and dryness as the most significant factors for its distribution.

We consider these works as a starting point for our efforts. However, we highlight several methodological differences: i) while they identified \textit{a priori} risk areas, we conduct a \textit{a posteriori} analysis on top of them; ii) we shift the paradigm from an unsupervised to a supervised one, modeling the presence of a vector as a binary classification problem; iii) we do not rely on hand-crafted features, as we leverage satellite images for predicting the target directly, allowing the convolutional model to learn the proper features for the task at hand.  

\subsection{Crop and Land classification}

In the remote sensing field, numerous researches focus on crop and land cover classification. A widely used approach relies on various indices, such as the Normalized Difference Vegetation Index (NDVI) and the Normalized Difference Water Index (NDWI). These two indices arise directly from raw spectral bands and reflect a clear physical meaning: while the former quantifies the green biomass, the latter depicts the water-body mapping. However, they show some limitations: as an example, the denser the vegetation coverage becomes, the more they will saturate, and the more performance will degrade~\cite{gao2015optical}. 

In such a case, the Enhanced Vegetation Index (EVI) represents a valid alternative:~\cite{zhong2018deep} used the EVI time-series as input for a deep learning architecture, aiming to classify summer crops (i.e., Rice, Corn, Safflower, etc.). In this respect, the motivations behind the adoption of time-series emerge from seasonal patterns in the dynamics of vegetation. Moreover, the authors showed that traditional machine learning approaches (XGBoost, Random Forest, and Support Vector Machine) lead to lower performance with respect to the proposed deep model, which exploits one-dimensional convolutions for capturing both seasonal patterns and small scale temporal variations. 

Differently, the authors of~\cite{ji20183d} described a three-dimensional Convolutional Neural Network (3D-CNN), inferring crops categories (i.e. Corn, Rice, Soyb and Tree) from satellite images acquired by the Gaofen-2 satellite~\cite{ming2015gaofen}. Since vegetation indices rely on few bands among the available ones, some crucial information could be discarded, thus resulting in lower generalization capabilities in the crop classification task. On this latter point, the authors propose the 3D-CNN, as it can look for temporal patterns and dynamics within the raw bands sequence directly.

Similarly, the authors of~\cite{zhang2017band} proposed a land cover classification method, with four different classes (crop, tree, water, and road), underpinning Sentinel bands as input to a Support Vector Machine (SVM) classifier. In this respect, they conduct experiments on three different types of features: vegetation indices, hand-selected bands, and, finally, all bands. As observed in~\cite{ji20183d}, the authors claimed that exploiting all bands leads to better results.

% Instead, in~\cite{bernasconi2019satellite}, the authors propose CNNs to extract suitable representations for land cover classification. They estimate the percentage of land occupied by a given category (i.e. lakes, industrial areas, vegetation, etc.). In doing so, they exploit a large labeled dataset of Sentinel 2 images, namely EuroSAT~\cite{helber2019eurosat}. It consists of 27'000 images labeled with ten different classes, each one of 64x64 pixel size. Eventually, they implemented a sliding-window approach to estimate land cover indicators from RGB images, the latter being bigger than the ones in the EuroSAT dataset.
%
\section{Datasets}
\label{sec:dataset}

\begin{figure}
    \centering
\begin{tabular}{@{\hspace{0cm}}c@{\hspace{0.2cm}}c}
    \includegraphics[width=0.45\columnwidth]{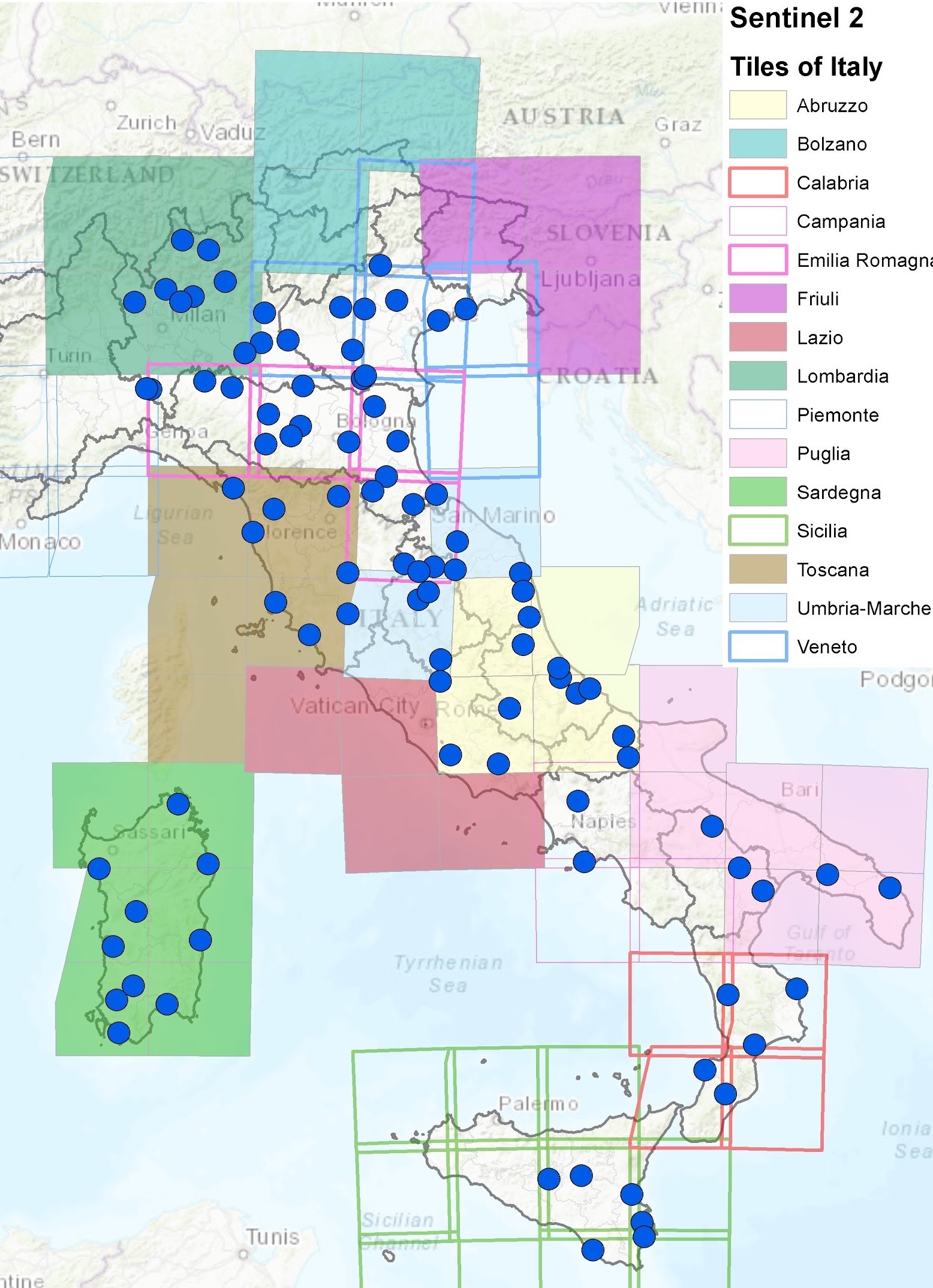} &
    \includegraphics[width=0.50\columnwidth]{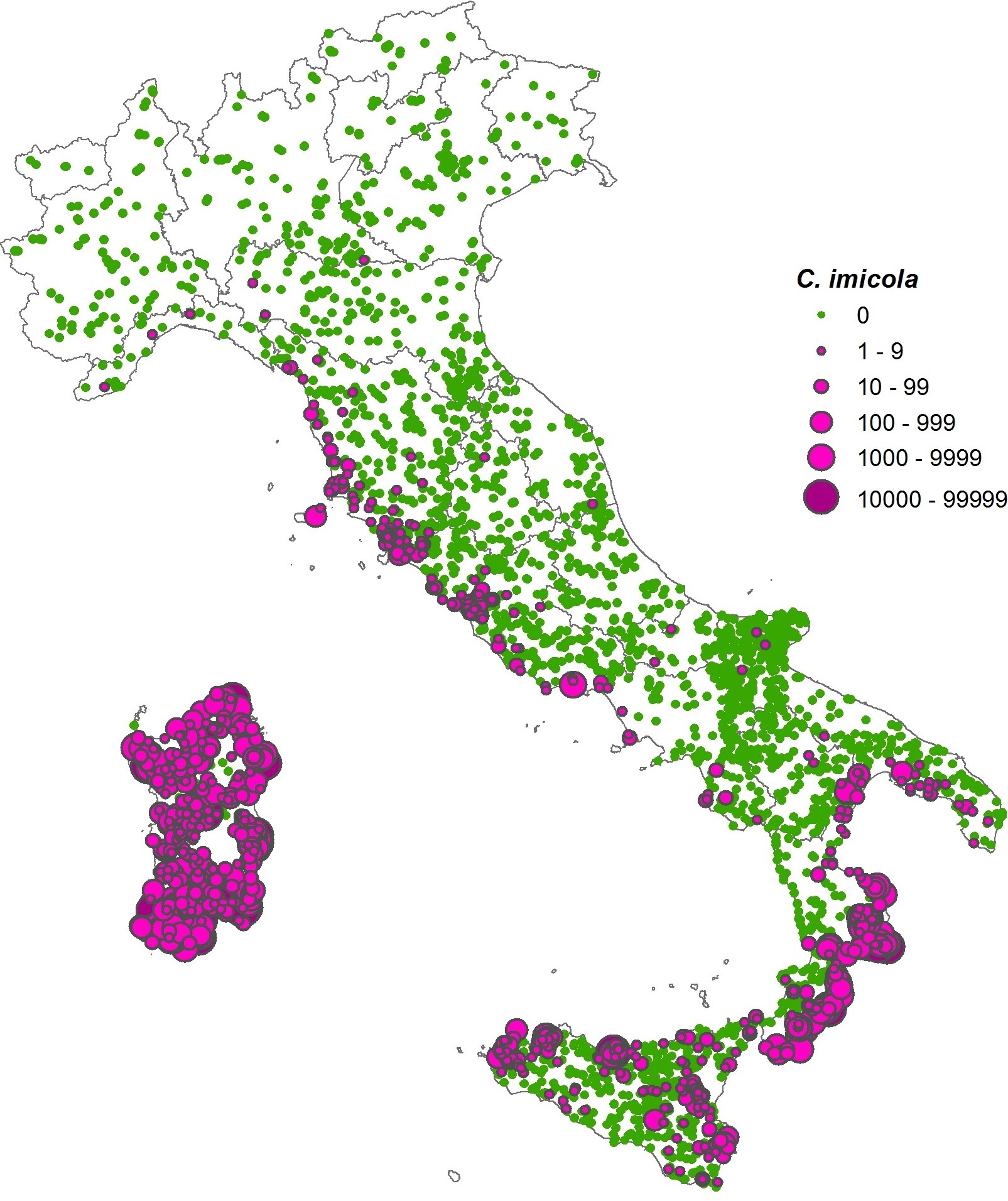} \\
\end{tabular}   
\caption{Left: Italian locations of 94 capture farms (2018). Right: spatial distribution of the \textit{C. imicola}, ranging from August 2000 to August 2017~\cite{ippoliti2019defining}.}
\label{fig:imicola_distr}
\end{figure}

\subsection{Bluetongue vectors Dataset}

Aiming to predict the presence of \textit{C. imicola}, we gathered a dataset coupling satellite patches depicting the Italian territory with ground-truth information regarding vectors' catches. 
In more details, we extracted the data regarding the presence/absence of the \textit{C. imicola} species from the `Culicoides collections'. Such information have been stored in a centralized database, hosted by the IZSAM Institute\footnote{Namely, Istituto Zooprofilattico Sperimentale dell'Abruzzo e del Molise.} as part of the Bluetongue National Surveillance Plan~\cite{goffredo2004entomological}. This plan takes place in Italy since 2000~\cite{giovannini2004surveillance}, and defines field and laboratory protocols for the collection of \textit{Culicoides} and their identification. 
The catches of these insects are conducted near the farms through various traps, activated in the evening and collected the next morning. Eventually, an entomological laboratory examines the traps for identifying the \textit{Culicoides} species and their actual amount.

\subsection{Satellite Imagery}

\begin{table}
\caption{Spectral bands from Sentinel-2 MSI sensor, coupled with their wavelength and resolutions.}
    \begin{center}
        \resizebox{\columnwidth}{!}{
        \begin{tabular}{l l c c}
        \toprule
        & \mytabheader{Bands} & \mytabheader{Wavelength ($\mu$m)} & \mytabheader{Res (m)} \vspace{0.1cm}\\
        \hline\addlinespace[0.1cm]
        $B_{1}$ & Coastal Aerosol & 0.443 & 60 \\
        $B_{2,3,4}$ & BGR channels & 0.490 & 10 \\
        $B_{5}$ & Vegetation Red Edge & 0.705 & 20 \\
        $B_{6}$ & Vegetation Red Edge & 0.740 & 20 \\
        $B_{7}$ & Vegetation Red Edge & 0.783 & 20 \\
        $B_{8}$ & NIR & 0.842 & 10 \\
        $B_{8\text{A}}$ & Vegetation Red Edge & 0.865 & 20 \\
        $B_{9}$ & Water Vapour & 0.945 & 60 \\
        $B_{10}$ & SWIR (Cirrus) & 1.375 & 60 \\
        $B_{11}$ & SWIR & 1.610 & 20 \\
        $B_{12}$ & SWIR & 2.190 & 20 \\
        \bottomrule
        \end{tabular}
    }
    \label{tab2}
    \end{center}
\end{table}

As mentioned before, the multi-spectral images we rely on are acquired through optical devices onboard the Sentinel-2A and 2B satellites. These images depict the Earth's surface with a high spatial, temporal, and spectral resolutions.

The two satellites simultaneously orbit around the Earth on the same trajectory, at the height of about 780 km, staggered between them by 180 degrees. They provide a multispectral "photograph" of the territory that they fly over every five days.
The spectral bands (see Table~\ref{tab2}) are acquired under different resolutions, in particular at 10, 20 and 60 meters per pixel. The ESA makes these images available through an online platform~\footnote{Please refer to \url{https://scihub.copernicus.eu} for additional details.}.

%In more details, the MSI sensor exploits the diverse response of the materials to light exposure. In fact, by collecting just a portion of the light rays, reflected in a certain range of the spectrum, they are able to detect specific features. 
%
In the following list, we sum up the main characteristics for each band, arranged in ascending order with respect to their wavelength:

\begin{itemize}
    \item \textbf{$B_1$}: it identifies any type of aerosol;
    \item \textbf{$B_2$}: the blue color, useful to discriminate between the different types of vegetation. Compared to lower frequencies, it provides a better clear water penetration and a superior illumination of the materials in shadow;
    \item \textbf{$B_3$}: the green color, which excels in highlighting muddy water, and helps in finding oil on the water surface and vegetation;
    \item \textbf{$B_4$}: the red color, suitable for finding urban and soil features; its good response to dead foliage allows to discriminate between the vegetation types;
    \item \textbf{$B_{5, 6, 7, 8A, 8}$}: all these bands are designed specifically for analyzing and classifying the vegetation. In particular, the last one enables shorelines mapping and extracting information about biomass content;
    \item \textbf{$B_{9, 10}$}: belonging to the infra-red region, they both aim at identifying water vapour. Notably, the second one is specific for cirrus clouds;
    \item \textbf{$B_{11, 12}$}: the last two bands distinguish areas covered by ice and snow from clouds.
\end{itemize}

\subsection{Datasets integration}

\begin{figure*}[t]
\centering
\includegraphics[width=0.95\textwidth]{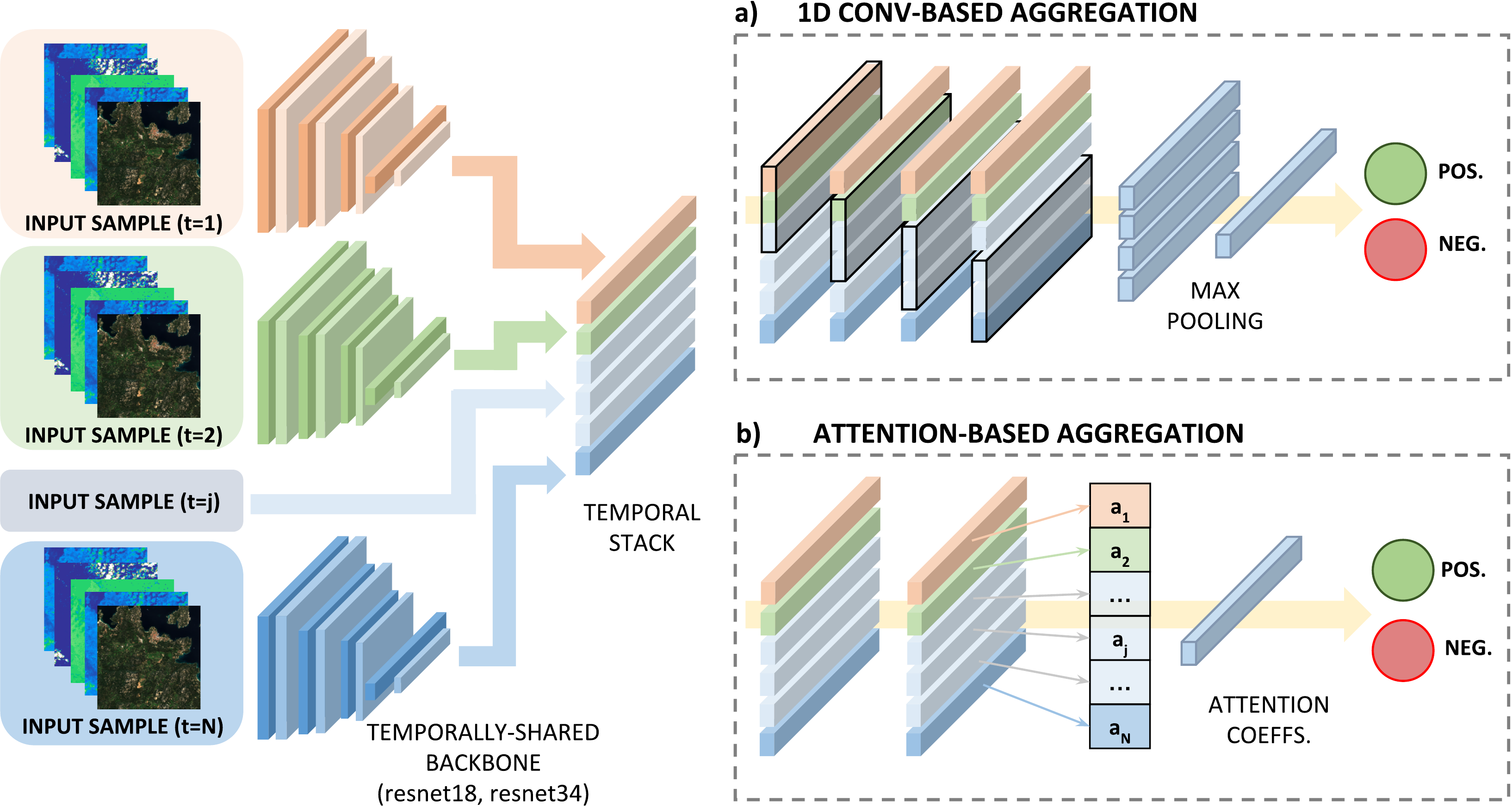}
\caption{The proposed framework, in its multi-instance variant: it takes multiple samples as input, each of which is independently processed by a shared CNN, the latter acting as the feature extractor (e.g. resnet18). Afterward, we stack those representations and form the input for the next neural aggregation module, which fulfills one of the following two options: (a) a 1D-Convolution based schema, leveraging the underlying time ordering; (b) an attention-based module, which estimates the importance of each representation during the overall computation. Best viewed in color.}
\label{fig:MIL}
\end{figure*}

\begin{table}
\caption{For each Italian region, the number of entomological catches and satellite images for the year 2018.}
\begin{minipage}{.50\columnwidth}
\resizebox{\columnwidth}{!}{
        \setlength{\tabcolsep}{0.25em}
        \begin{tabular}{l c c c}
        \toprule
        \mytabheader{Regions} & \mytabheader{Farms} & \mytabheader{Catches} & \mytabheader{Images}  \vspace{0.1cm}\\
        \hline\addlinespace[0.1cm]
        Abruzzo & 6 & 196 & 491  \\
        Basilicata & 2 & 102 & 142  \\
        Calabria & 5 & 192 & 495  \\
        E. Romagna & 13 & 489 & 999  \\
        Lazio & 3 & 112 & 216  \\
        Lombardy & 13 & 431 & 984  \\
        Marche & 5 & 229 & 429  \\
        \bottomrule
        \end{tabular}
}
\end{minipage}
\begin{minipage}{.46\columnwidth}
\resizebox{\columnwidth}{!}{
        \setlength{\tabcolsep}{0.25em}
        \begin{tabular}{l c c c}
        \toprule
        \mytabheader{Regions} & \mytabheader{Farms} & \mytabheader{Catches} & \mytabheader{Images} \vspace{0.1cm}\\
        \hline\addlinespace[0.1cm]
        Molise & 1 & 26 & 69  \\
        Apulia & 4 & 196 & 280  \\
        Sardinia & 10 & 419 & 778  \\
        Sicily & 6 & 188 & 423 \\
        Tuscany & 8 & 374 & 702 \\
        Veneto & 10 & 384 & 792 \\
        Umbria & 6 & 262 & 503 \\
        \bottomrule
        \end{tabular}
}
\end{minipage}
\label{tab:regions_dataset}
\end{table}

We combined the ground-truth information regarding \textit{C. imicola} catches with the ones coming from Sentinel-2 satellites. With no loss of generality, we restrict our analysis to the year 2018, as Sentinel-2B data have been made available since March 2017 only. 

For each capture, we collect the corresponding satellite image depicting its site location. To avoid misleading associations between input and targets, we focus on farms with an accurate georeferencing only. Importantly, we create each example so that the image's acquisition date is as close as possible to the time at which the trap was placed. Among the different spatial resolutions available, we adopt 20 meters as the default input resolution. We empirically observe that it represents a good compromise between preserving images' details and the overall memory footprint. This way, for each capture site, we obtain a 224x224 pixel image (resulting in a cutout of 4480x4480 $m$), which captures environmental information near the farm. These catches take place periodically, at variable intervals.

Given a capture, the presence or absence of the \textit{C. imicola} represents the ground truth we adopt during our experiments. The dataset is unbalanced in favor of the absence of the vector. 
On the one hand, the positive catches constitute approximately one-seventh of the total. On the other hand, the distribution of positive labels is not uniform among different regions: in this respect, Figure~\ref{fig:imicola_distr} (right) shows the distribution of the catches over the Italian territory, as well as the tiles of the satellite images (left). 

To sum up, the final dimension of the dataset equals the number of catches, in our case 3671, distributed among 94 farms. For our purposes, we leverage 7514 satellite images. As mentioned before, 3182 samples refer to the negative class, thus indicating the absence of the \textit{C. imicola}, while 489 samples belong to the positive one. We propose in Table~\ref{tab:regions_dataset} a region-wise summary regarding the dataset.

\section{Models}
\label{sec:models}

\subsection{Features Extraction}

Our proposal heavily relies on ResNet~\cite{he2016deep} as feature extractor for the overall architecture. In more details, ResNet is one of the most used networks for classification and object detection tasks.
The underlying idea - which makes this architecture so powerful - regards the residual blocks, which aim to simplify the training of neural networks characterized by many layers. In this regards, several researches~\cite{simonyan2014very, szegedy2015going} highlight networks' depth as a crucial factor: indeed, deeper models allow to learn richer representations. On the other hand, the deeper the networks become, the more the vanishing gradient issue~\cite{glorot2010understanding} emerges.
A careful initialization has addressed this problem primarily~\cite{glorot2010understanding}, as well as placing intermediate normalization layers~\cite{ioffe2015batch}. Differently, ResNet and its variants tackle the degradation problem by means of skip connections, as the network is equipped with an identity mapping between each block and the following one. This way, the gradients can be back-propagated without suffering from numerical issues. Moreover, identity shortcut connections add neither additional parameters nor computational complexity. 

Since we rely on a few labeled data, our proposal exploits a ResNet pre-trained model on the Imagenet dataset~\cite{russakovsky2015imagenet}, and then fine-tune it on ours. Since training a model from scratch could be hard, a standard guideline envisions a pre-training stage on a large dataset. Therefore, one could take such an achieved optimum as an initialization point for the next optimization phase, the latter guided by the task at hand.
% As an alternative, in the case of very few data, the features extractor paradigm suggests updating solely the final layers, the latter responsible for computing predictions. The underlying idea stands on the fact that low-level layers provide raw features, the ones that can be potentially shared among different tasks. Such a scheme does not fit on last layers, as they become specific towards the targets defined for the original task.

\subsection{Attention-based Features Aggregation}
\label{attention}

As we rely - for both the training and the inference stages - on multiple images at a time, we designed two solutions for aggregating representations coming from each sample. Let us start from the first one: we discard information arising from the temporal order and focus on features of each image solely.

Such an approach is based on the work described in~\cite{ilse2018attention}, where the authors proposed a novel neural module for handling Multiple Instance Learning (MIL) problems. MIL represents a variation of the supervised learning paradigm, where a single label refers to a bag of instances. Considering $X = \{x_1, x_2, \ldots, x_K\}$ as a bag of samples exhibiting neither dependency nor ordering among each other, the labels of each individual sample are $y_{1}, y_{2}, \ldots, y_{K}$, with $y_{k} \in \{0,1\}$; however, they remain unknown during the training phase. In this respect, the assumption underlining the MIL framework can be summarized as follows:

\begin{equation}
Y = 
\begin{cases} 
0 ,& \text{iff } \sum_k y_{k} = 0, \\
1 ,&\text{otherwise} .
\end{cases}
\label{eq_1}
\end{equation}

The authors' proposal consists in training a MIL model by optimizing the log-likelihood function. Importantly, the probability assigned to a given bag must be permutation invariant: this way, the solution to the MIL problem arises naturally from the Fundamental Theorem of Symmetric Functions~\cite{zaheer2017deep}. 

We inherit from~\cite{ilse2018attention} a general three-steps approach for classifying a bag of instances: i) to apply a non-linear transformation to instances trough a function $f$; ii) to exploit a symmetric function $\sigma$ to aggregate these features; iii) to transform such a comprehensive feature vector again by means of a function $g$. 
In our work, we adopt the embedding level approach for designing the function $f$, the latter mapping instances to a low dimensional embedding through a deep neural network. This way, the whole approach results flexible and trainable end-to-end. 
For the second step - the one implementing the aggregation phase - we rely on the attention mechanism. In this regards, we compute a weighted average of instances' latent representations, where a multi-layered perceptron outputs the coefficients. Given a bag of k embeddings $H = \{h_1, \ldots, h_K\}$ and parameters $W$ and $V$, we combine them as follows:

\begin{equation}\label{eq:weighted_sum}
z = \sum_{k=1}^{K} a_k h_k,
\end{equation}
where:
\begin{equation}\label{eq:attention}
a_{k} = \dfrac{ \exp\{W^{\top} \tanh \big{(} V h_{k}^{\top} \big{)}\}}{{\displaystyle \sum_{j=1}^K} \exp \{ W^{\top} \tanh \big{(}V h_{j}^{\top}\big{)} \} }.
\end{equation}

Our assumptions deviate from the MIL framework ones, where the bag would be labeled as positive whether at least one of its sample is positive (see Eq.~\ref{eq_1}). In our problem, differently, we evaluate how past images influence the prediction for the current one, regardless of their order.

\subsection{Convolution-based Features Aggregation}
\label{Conv1d}

As a second step, we aimed to assess whether the temporal information could bring meaningful features for the task at hand. In this regards, we compare the above-described proposal with a different one, which exploits the images' acquisition time. Starting from a query image, this alternative takes as input five images from the temporal neighborhood of the query, restricting to the ones acquired before. We then arrange the images according to their temporal order. Afterward, the ResNet backbone independently processes each sample within the sequence, so resulting in a list of 512-dimensional embeddings. We then apply a one-dimensional convolution, followed by a max-pooling stage, on the temporal axis (see Fig.~\ref{fig:MIL} a)). Eventually, a linear layer outputs the binary predictions (the same holds for the attention-based architecture).

During the preliminary stage, we intentionally discard recurrent networks for designing such a time-aware approach. Indeed, we would not benefit from the capabilities of these networks, as we focus on short and fixed-length input sequence. In support of this,~\cite{bai2018empirical} presented an empirical evaluation of convolutional and recurrent architectures. From their remarks, convolution arises as a natural starting point for sequence modeling problems; this is due to its low memory requirements, its stable gradient and its suitability for parallelism. Eventually, as shown in~\cite{zhong2018deep}, the one-dimensional convolutions are able to extract features from satellite imagery, thus being viable for classifying crops.

\section{Experiments}
\label{sec:experiments}

\subsection{Evaluation protocol}
\label{sec:protocol}

For avoiding overfitting and misleading conclusions, we adopt the Stratified K-Fold cross-validation to evaluate the performance of our models (with $\text{K}=5$). Stratified K-Fold is defined as a simple variation of the K-fold, in a way that each fold preserves approximately the probability distribution of the classes (the one that can be computed on the complete dataset).
Moreover, before splitting the dataset, we make sure each fold holds samples from all regions under consideration; thus guaranteeing the model to inspect images with different environmental and vegetative characteristics.

\subsection{Metrics}

We propose several metrics to evaluate the performance among various scenarios and architectures. Since the dataset at hand is actually unbalanced, an high accuracy value does not represent a symptom of satisfying performance nor generalization capabilities. Differently, metrics as precision and recall are suitable, as well as the F1-score, the latter rewarding solutions showing high values for both of the former. Eventually, we assess the models computing the Area Under the Precision-Recall Curve (AUPRC) for the positive class, thus obtaining a threshold-invariant summary.

\subsection{Implementation details}

Before being fed into the network, we normalize each spectral band independently, by computing the 2\textsuperscript{nd} and 98\textsuperscript{th} percentile values. We use these ones as boundaries for clipping each pixel. We empirically found that percentiles gain robustness when compared to the minimum and maximum, which suffer in presence of noisy acquisitions. After clipping the pixels' values accordingly, we apply the min-max normalization technique~\cite{prathap2018deep}.

During each experiment, we exploit the pre-trained weights for finetuning ResNet. We apply data augmentation to the input images, as horizontal flip, vertical flip, and rotation. In our experiments, we train the model for 150 epochs, setting the learning rate fixed at 0.001. The batch size equals to 16. For regularization purpose, we apply the dropout technique, with a drop probability of 0.2.

For mitigating the above-discussed classes unbalance, we exploit the weighted counterpart of the cross-entropy loss. In so doing, the weight of the positive class is seven times greater the weight of the negative one.

\subsection{Baselines}

Firstly, we consider as baseline classifier the one implementing a naive strategy, as generating random predictions by sampling from the class distribution of the training set. Indeed, such a naive choice leads to a measure of `baseline' performance, namely the expected success rate when simply guessing. This way, we are able to evaluate whether the model effectively learns from the data.

Secondly, we assess a single-instance binary classifier, equipped with resnet18 as feature extractor. Moreover, we tested it on different input settings, such as RGB, the Normalized Difference Vegetation Index (NDVI) coupled with the Normalized Difference Water Index (NDWI) (see Eq.~\ref{eqn:NDVI_NDWI}), and all 13 bands. Moreover - as previously done in~\cite{zhang2017band} - we test the $B_{4}$, $B_{8A}$ and $B_{11}$ bands, the ones involved during the NDVI and NDWI computations. 
The reasons behind these analyses are twofold: on the one hand, we can evaluate whether more acquisition will lead to significant improvements; on the other hand, we get a clear view on the contribution of spectral bands when compared to their hand-designed derivatives.

\begin{equation}
\text{NDVI} = \frac{B_{8A}-B_{4}}{B_{8A}+B_{4}}; \quad \text{NDWI} = \frac{B_{8A}-B_{11}}{B_{8A}+B_{11}}. \\
\label{eqn:NDVI_NDWI}
\end{equation}
\vspace{0.1cm}

\subsection{Single-instance Classification}

\begin{table}[t]
\caption{Results obtained in the single-instance setting.}
    \begin{center}
        \resizebox{\columnwidth}{!}{
        \setlength{\tabcolsep}{0.5em}
        \begin{tabular}{l c c c c c c}
        \toprule
         \mytabheader{Model [pre-trained yes/no]} & \mytabheader{Acc.} & \mytabheader{Pr.} & \mytabheader{Rc.} & \mytabheader{F1} & \mytabheader{AUPRC} \vspace{0.1cm}\\
         \hline\addlinespace[0.05cm]
         Random classifier [\xmark] & .819 & .169 & .140 & .153 & .116\\
         \hline\addlinespace[0.05cm]
         \hline\addlinespace[0.1cm]
         RGB [\xmark] & .922 & .702 & .584 & .637 & .606\\
         RGB [\cmark] & .931 & .764 & .616 & .679 & .653\\
         \hline\addlinespace[0.1cm]
         NDVI + NDWI [\xmark] & .926 & .708 & .640 & .668 & .656\\
         NDVI + NDWI [\cmark] & .928 & .714 & .616 & .659 & .668\\
         \hline\addlinespace[0.1cm]
         $B_{4}$ + $B_{8A}$ + $B_{11}$ [\xmark] & .908 & .612 & .607 & .609 & .632\\
         $B_{4}$ + $B_{8A}$ + $B_{11}$ [\cmark] & .912 & .697 & .602 & .643 & .641\\
         \hline\addlinespace[0.1cm]
         Spectral bands [\xmark] & .913 & .619 & .705 & .656 & .668\\
         Spectral bands [\cmark] & .924 & .667 & .721 & \textbf{.689} & \textbf{.682}\\
         \bottomrule
        \end{tabular}
    }
    \label{tab:tab3}
    \end{center}
\end{table}

We report in Table~\ref{tab:tab3} the results for the single-instance setting. Looking at the first row, the one showing the random classifier's performance, we can draw the following considerations: firstly, the problem unbalance makes the accuracy unsuitable for judging the model's capabilities; secondly, as all variants achieve higher F1 and AUPRC scores, our model effectively learns useful patterns from the data.

Looking at Table~\ref{tab:tab3}, we draw the following remarks:

\begin{itemize}
    \item The raw bands lead to the highest performance for our problem. Such a result strengths the underlying intuition of our work, which exploits the environmental patterns arising from raw satellite images for predicting the presence of \textit{C. imicola};
    \item The NDVI and NDWI indices yield better results than the $B_{4}$, $B_{8A}$ and $B_{11}$ bands, namely the ones from which the former are computed: such a finding highlights their relevance for the field. However, it is worth noting that exploiting all the channels brings to even better results; thus indicating the occurrence of significant features ignored by the NDVI and NDWI;
    \item We remark the beneficial effects arising from the ImageNet pre-training: despite the ours being a completely different task, the convolutional weights can be inherited and fine-tuned for working on spectral bands. This may be useful when facing remote sensing applications, especially the ones characterised by few labeled data. 
\end{itemize}

For additional notes on the single-istance setting, please refer to Subsection~\ref{sec:limitations} .

\subsection{Multi-instance Classification}

\begin{table}[t]
\caption{Results obtained in the multi-instance setting.}
    \begin{center}
    \resizebox{\columnwidth}{!}{
        \setlength{\tabcolsep}{0.5em}
        \begin{tabular}{l c c c c c}
        \toprule
        \mytabheader{Model} & \mytabheader{Acc.} & \mytabheader{Pr.} & \mytabheader{Rc.} & \mytabheader{F1} & \mytabheader{AUPRC} \vspace{0.1cm}\\
        \hline\addlinespace[0.05cm]
         Single-istance & .924 & .667 & .721 & .689 & .682\\
         \hline\addlinespace[0.05cm]
         \hline\addlinespace[0.1cm]
         Mean-based & .957 & .783 & .877 & \textbf{.826} & .804\\
         \hline\addlinespace[0.1cm]
         Attention-based & .955 & .786 & .853 & .817 & .812\\
         \hline\addlinespace[0.1cm]
         Conv-based & .956 & .786 & .860 & .821 & \textbf{.821}\\
        \bottomrule
        \end{tabular}
    }
    \label{tab4}
    \end{center}
\end{table}

In the multiple-instance setting we evaluate three variants: i) Attention-based (Section~\ref{attention}); ii) 1D-Convolution (Section~\ref{Conv1d}); iii) a Mean-based approach, the latter computing the mean of ResNet's activations along the temporal dimension. We design such a baseline for investigating the impact of the proposed aggregation modules w.r.t. a naive one. Indeed, averaging the representations does not consider the samples' ordering, nor it is able to identify the key instances.
It is worth noting that the three variants compared in this section takes all spectral bands as input. We apply the same data augmentation and hyperparameters adopted per the single-instance setting, except for the number of epochs, which is lowered to 120 due to faster convergence.

We report the results in Table~\ref{tab4} and observe that:
\begin{itemize}
    \item The single-instance classifier earns worse results w.r.t. the multi-instance one. Indeed, the latter can capture temporal patterns, appearing in more than one image. Additionally, in the single-instance approach, the prediction could suffer in the presence of noisy acquisitions. As an example, we mention the cloudiness, which affects a large part of the available data. Differently, under the multi-instance setting, it is unlikely that all five images are cloudy;
    \item The Conv-based model leads to slightly better results w.r.t. the attention-based one; thus indicating that the order contains some valuable information for the task;
    \item Both the Attention-based and the Conv-based models exhibit a better AUPRC value compared to the Mean-based one. Nevertheless, F1-score does not support this result. We will conduct further investigations to assess which of the two metrics is more suitable for the problem at hand.
\end{itemize}

\subsection{Limitations and Future Works}
\label{sec:limitations}

Even though the results being suitable, we found that the spectral bands perform just slightly better than the RGB ones (especially when looking at the F1-score, see Table~\ref{tab:tab3}). This indicates the \textit{curse of dimensionality} when dealing with complex data as the multi-spectral images. In future works, we will explore this aspect, trying to increase the gap between spectral bands and other hand-crafted representations. In this regards - further than gathering more data - we envision a network pre-training on crop and land classification directly, for which large datasets~\cite{sumbul2019bigearthnet} are available currently. This could result in a proper initialization, thus improving the generalization capabilities of our model. 

As discussed in Subsection~\ref{sec:protocol}, we assess our proposal through Stratified K-Fold. In doing so, each of the K runs has been conducted by splitting entomological information about catches in two non-overlapping sets (namely, the training and test set). As a consequence, the same capture site could appear in both the training and test set, even though under different time periods. Because of this, we cannot claim for sure whether the model generalizes to unseen sites. We will remove such a limitation in future works, defining a clear cross-site evaluation protocol.

As can be appreciated in Fig.~\ref{fig:imicola_distr} (right), the \textit{C. imicola} shows some subtle patterns in its geographical distribution. In particular, it follows a sort of locality principle, stating that neighboring sites are likely to share the same levels of abundance. Thus, one could infer the presence of the \textit{C. imicola} in a certain area just looking to its neighboring nodes, modeling the overall environment as a graph connecting sites and regions. This way, we could frame the problem under the graph-based semi-supervised framework~\cite{zhu2003semi}, the latter assuming the label information to be smoothed over the graph. In this respect, we will profit from recent advances on Graph Convolutional Neural Networks (GCNNs)~\cite{kipf2016semi,porrello2019classifying} and propose a novel inference engine, which would exploit the local dynamics arising from the spread of \textit{C. imicola}.

\section{Conclusion}
\label{sec:conclusion}

In this work, we focus on predicting whether the \textit{C. imicola} vector is present in the area of interest. We achieve that through a novel deep-learning-based approach, which extracts meaningful features from a sequence of satellite imagery depicting that area. Even though a bigger dataset may enhance its performance, our model already provides suitable generalization capabilities than both the usual hand-crafted features and the RGB bands only; therefore, all the spectral bands must carry fundamental features about the spread of the vector. Furthermore, we propose a multi-instance model leveraging temporal patterns, which leads to better results than single-instance one. Future work aims to improve the overall performance through a targeted pre-training phase, over a large dataset specific for satellite imagery. Moreover, as we observe locality playing a significant role in the geographical spreading of \textit{C. imicola}, we will frame the here-discussed problem in terms of semi-supervised learning over graphs.

\section*{Acknowledgment}
This work was done within the UNIMORE project `AI4V'. Funding was provided by the Italian Ministry of Health - \url{www.salute.gov.it} - (IZSAM\_01/18\_RC, Current Research 2018 “Artificial intelligence and remote sensing: innovative methods for monitoring the vectors and the associated ecological/environmental variables”). 
{            
  \bibliographystyle{IEEEtran}
  \bibliography{IEEEabrv,mybib}

\begin{thebibliography}{10}
\providecommand{\url}[1]{#1}
\csname url@samestyle\endcsname
\providecommand{\newblock}{\relax}
\providecommand{\bibinfo}[2]{#2}
\providecommand{\BIBentrySTDinterwordspacing}{\spaceskip=0pt\relax}
\providecommand{\BIBentryALTinterwordstretchfactor}{4}
\providecommand{\BIBentryALTinterwordspacing}{\spaceskip=\fontdimen2\font plus
\BIBentryALTinterwordstretchfactor\fontdimen3\font minus
  \fontdimen4\font\relax}
\providecommand{\BIBforeignlanguage}[2]{{%
\expandafter\ifx\csname l@#1\endcsname\relax
\typeout{** WARNING: IEEEtran.bst: No hyphenation pattern has been}%
\typeout{** loaded for the language `#1'. Using the pattern for}%
\typeout{** the default language instead.}%
\else
\language=\csname l@#1\endcsname
\fi
#2}}
\providecommand{\BIBdecl}{\relax}
\BIBdecl

\bibitem{semenza2017vector}
J.~C. Semenza and J.~E. Suk, ``Vector-borne diseases and climate change: a
  european perspective,'' \emph{FEMS microbiology letters}, vol. 365, no.~2, p.
  fnx244, 2017.

\bibitem{conte2007influence}
A.~Conte, M.~Goffredo, C.~Ippoliti, and R.~Meiswinkel, ``Influence of biotic
  and abiotic factors on the distribution and abundance of culicoides imicola
  and the obsoletus complex in italy,'' \emph{Veterinary parasitology}, vol.
  150, no.~4, pp. 333--344, 2007.

\bibitem{mcintyre2017systematic}
K.~M. McIntyre, C.~Setzkorn, P.~J. Hepworth, S.~Morand, A.~P. Morse, and
  M.~Baylis, ``Systematic assessment of the climate sensitivity of important
  human and domestic animals pathogens in europe,'' \emph{Scientific reports},
  vol.~7, no.~1, p. 7134, 2017.

\bibitem{drusch2012sentinel}
M.~Drusch, U.~Del~Bello, S.~Carlier, O.~Colin, V.~Fernandez, F.~Gascon,
  B.~Hoersch, C.~Isola, P.~Laberinti, P.~Martimort \emph{et~al.}, ``Sentinel-2:
  Esa's optical high-resolution mission for gmes operational services,''
  \emph{Remote sensing of Environment}, vol. 120, pp. 25--36, 2012.

\bibitem{berger2012esa}
M.~Berger, J.~Moreno, J.~A. Johannessen, P.~F. Levelt, and R.~F. Hanssen,
  ``Esa's sentinel missions in support of earth system science,'' \emph{Remote
  Sensing of Environment}, vol. 120, pp. 84--90, 2012.

\bibitem{sutskever2012imagenet}
I.~Sutskever, G.~E. Hinton, and A.~Krizhevsky, ``Imagenet classification with
  deep convolutional neural networks,'' \emph{Advances in Neural Information
  Processing Systems}, pp. 1097--1105, 2012.

\bibitem{he2016deep}
K.~He, X.~Zhang, S.~Ren, and J.~Sun, ``Deep residual learning for image
  recognition,'' in \emph{IEEE International Conference on Computer Vision and
  Pattern Recognition}, 2016, pp. 770--778.

\bibitem{ippoliti2019defining}
C.~Ippoliti, L.~Candeloro, M.~Gilbert, M.~Goffredo, G.~Mancini, G.~Curci,
  S.~Falasca, S.~Tora, A.~Di~Lorenzo, M.~Quaglia \emph{et~al.}, ``Defining
  ecological regions in italy based on a multivariate clustering approach: A
  first step towards a targeted vector borne disease surveillance,'' \emph{PloS
  one}, vol.~14, no.~7, p. e0219072, 2019.

\bibitem{bailey2004identifying}
R.~G. Bailey, ``Identifying ecoregion boundaries,'' \emph{Environmental
  management}, vol.~34, no.~1, pp. S14--S26, 2004.

\bibitem{ducheyne2013abundance}
E.~Ducheyne, M.~A.~M. Chueca, J.~Lucientes, C.~Calvete, R.~Estrada, G.-J.
  Boender, E.~Goossens, E.~M. De~Clercq, and G.~Hendrickx, ``Abundance
  modelling of invasive and indigenous culicoides species in spain,''
  \emph{Geospatial Health}, pp. 241--254, 2013.

\bibitem{gao2015optical}
Y.~Gao, J.~P. Walker, M.~Allahmoradi, A.~Monerris, D.~Ryu, and T.~J. Jackson,
  ``Optical sensing of vegetation water content: A synthesis study,''
  \emph{IEEE Journal of Selected Topics in Applied Earth Observations and
  Remote Sensing}, vol.~8, no.~4, pp. 1456--1464, 2015.

\bibitem{zhong2018deep}
L.~Zhong, L.~Hu, and H.~Zhou, ``Deep learning based multi-temporal crop
  classification,'' \emph{Remote Sensing of Environment}, vol. 221, pp.
  430--443, 12 2018.

\bibitem{ji20183d}
S.~Ji, Z.~Chi, A.~Xu, Y.~Shi, and Y.~Duan, ``3d convolutional neural networks
  for crop classification with multi-temporal remote sensing images,''
  \emph{Remote Sensing}, vol.~10, p.~75, 01 2018.

\bibitem{ming2015gaofen}
H.~G. H. L. J.~G. Ming~Li, Teng~Pan, ``Gaofen-2 mission introduction and
  characteristics,'' in \emph{Proceedings of the 66th International
  Astronautical Congress (IAC)}, 2015.

\bibitem{zhang2017band}
T.~Zhang, J.~Su, C.~Liu, W.-H. Chen, H.~Liu, and G.~Liu, ``Band selection in
  sentinel-2 satellite for agriculture applications,'' in \emph{2017 23rd
  International Conference on Automation and Computing (ICAC)}.\hskip 1em plus
  0.5em minus 0.4em\relax IEEE, 2017, pp. 1--6.

\bibitem{goffredo2004entomological}
M.~Goffredo and R.~Meiswinkel, ``Entomological surveillance of bluetongue in
  italy: methods of capture, catch analysis and identification of culicoides
  biting midges,'' \emph{Vet Ital}, vol.~40, no.~3, pp. 260--265, 2004.

\bibitem{giovannini2004surveillance}
A.~Giovannini, C.~Paladini, P.~Calistri, A.~Conte, P.~Colangeli, U.~Santucci,
  D.~Nannini, and V.~Caporale, ``Surveillance system of bluetongue in italy,''
  \emph{Vet Ital}, vol.~40, no.~3, pp. 369--84, 2004.

\bibitem{simonyan2014very}
K.~Simonyan and A.~Zisserman, ``Very deep convolutional networks for
  large-scale image recognition,'' \emph{arXiv preprint arXiv:1409.1556}, 2014.

\bibitem{szegedy2015going}
C.~Szegedy, W.~Liu, Y.~Jia, P.~Sermanet, S.~Reed, D.~Anguelov, D.~Erhan,
  V.~Vanhoucke, and A.~Rabinovich, ``Going deeper with convolutions,'' in
  \emph{IEEE International Conference on Computer Vision and Pattern
  Recognition}, 2015, pp. 1--9.

\bibitem{glorot2010understanding}
X.~Glorot and Y.~Bengio, ``Understanding the difficulty of training deep
  feedforward neural networks,'' in \emph{International Conference on
  Artificial Intelligence and Statistics}, 2010, pp. 249--256.

\bibitem{ioffe2015batch}
S.~Ioffe and C.~Szegedy, ``Batch normalization: Accelerating deep network
  training by reducing internal covariate shift,'' \emph{arXiv preprint
  arXiv:1502.03167}, 2015.

\bibitem{russakovsky2015imagenet}
O.~Russakovsky, J.~Deng, H.~Su, J.~Krause, S.~Satheesh, S.~Ma, Z.~Huang,
  A.~Karpathy, A.~Khosla, M.~Bernstein \emph{et~al.}, ``Imagenet large scale
  visual recognition challenge,'' \emph{International Journal of Computer
  Vision}, vol. 115, no.~3, pp. 211--252, 2015.

\bibitem{ilse2018attention}
M.~Ilse, J.~M. Tomczak, and M.~Welling, ``Attention-based deep multiple
  instance learning,'' \emph{arXiv preprint arXiv:1802.04712}, 2018.

\bibitem{zaheer2017deep}
M.~Zaheer, S.~Kottur, S.~Ravanbakhsh, B.~Poczos, R.~R. Salakhutdinov, and A.~J.
  Smola, ``Deep sets,'' in \emph{Neural Information Processing Systems}, 2017,
  pp. 3391--3401.

\bibitem{bai2018empirical}
S.~Bai, J.~Z. Kolter, and V.~Koltun, ``An empirical evaluation of generic
  convolutional and recurrent networks for sequence modeling,'' \emph{arXiv
  preprint arXiv:1803.01271}, 2018.

\bibitem{prathap2018deep}
G.~Prathap and I.~Afanasyev, ``Deep learning approach for building detection in
  satellite multispectral imagery,'' in \emph{2018 International Conference on
  Intelligent Systems (IS)}.\hskip 1em plus 0.5em minus 0.4em\relax IEEE, 2018,
  pp. 461--465.

\bibitem{sumbul2019bigearthnet}
G.~Sumbul, M.~Charfuelan, B.~Demir, and V.~Markl, ``Bigearthnet: A large-scale
  benchmark archive for remote sensing image understanding,'' \emph{arXiv
  preprint arXiv:1902.06148}, 2019.

\bibitem{zhu2003semi}
X.~Zhu, Z.~Ghahramani, and J.~D. Lafferty, ``Semi-supervised learning using
  gaussian fields and harmonic functions,'' in \emph{International Conference
  on Machine Learning}, 2003, pp. 912--919.

\bibitem{kipf2016semi}
T.~N. Kipf and M.~Welling, ``Semi-supervised classification with graph
  convolutional networks,'' \emph{arXiv preprint arXiv:1609.02907}, 2016.

\bibitem{porrello2019classifying}
A.~Porrello, D.~Abati, S.~Calderara, and R.~Cucchiara, ``Classifying signals on
  irregular domains via convolutional cluster pooling,'' in \emph{International
  Conference on Artificial Intelligence and Statistics}, 2019, pp. 1388--1397.

\end{thebibliography}
}  

\end{document}